\pdfoutput=1

\documentclass[11pt]{article}

\usepackage{emnlp2023}

\usepackage{times}
\usepackage{latexsym}
\usepackage{CJKutf8}
\usepackage{graphicx}
\usepackage{amsmath}
\usepackage{multirow}
\usepackage{bbding}
\usepackage{arydshln} 
\usepackage[T1]{fontenc}

\usepackage[utf8]{inputenc}

\usepackage{microtype}
\usepackage{tablefootnote}

\usepackage{inconsolata}
\newcommand{\tabincell}[2]{\begin{tabular}{@{}#1@{}}#2\end{tabular}}  

\title{CSCD-NS: a Chinese Spelling Check Dataset for Native Speakers}

\author{Yong Hu, Fandong Meng, Jie Zhou \\ 
WeChat AI, Tencent Inc., China \\
\texttt{\{rightyonghu,fandongmeng,withtomzhou\}@tencent.com}
}

\begin{document}
\maketitle
\begin{abstract}
In this paper, we present CSCD-NS, the first Chinese spelling check (CSC) dataset designed for native speakers, containing 40,000 samples from a Chinese social platform. Compared with existing CSC datasets aimed at Chinese learners, CSCD-NS is ten times larger in scale and exhibits a distinct error distribution, with a significantly higher proportion of word-level errors. To further enhance the data resource, we propose a novel method that simulates the input process through an input method, generating large-scale and high-quality pseudo data that closely resembles the actual error distribution and outperforms existing methods. Moreover, we investigate the performance of various models in this scenario, including large language models (LLMs), such as ChatGPT. The result indicates that generative models underperform BERT-like classification models due to strict length and pronunciation constraints. The high prevalence of word-level errors also makes CSC for native speakers challenging enough, leaving substantial room for improvement.
\footnote{https://github.com/nghuyong/cscd-ns}
\end{abstract}

\section{Introduction}
Chinese spelling check (CSC) is a task to detect and correct spelling errors in Chinese texts. 
There are two primary user groups for CSC: (1) Chinese learners, including teenage students and individuals who use Chinese as a second language, and (2) Chinese native speakers. It is obvious that the latter user group has a larger population and more diverse applications, therefore, this paper concentrates on CSC for native speakers.

\begin{CJK*}{UTF8}{gbsn}
\begin{figure}[t!]
  \centering
  \includegraphics[width=0.5\textwidth]{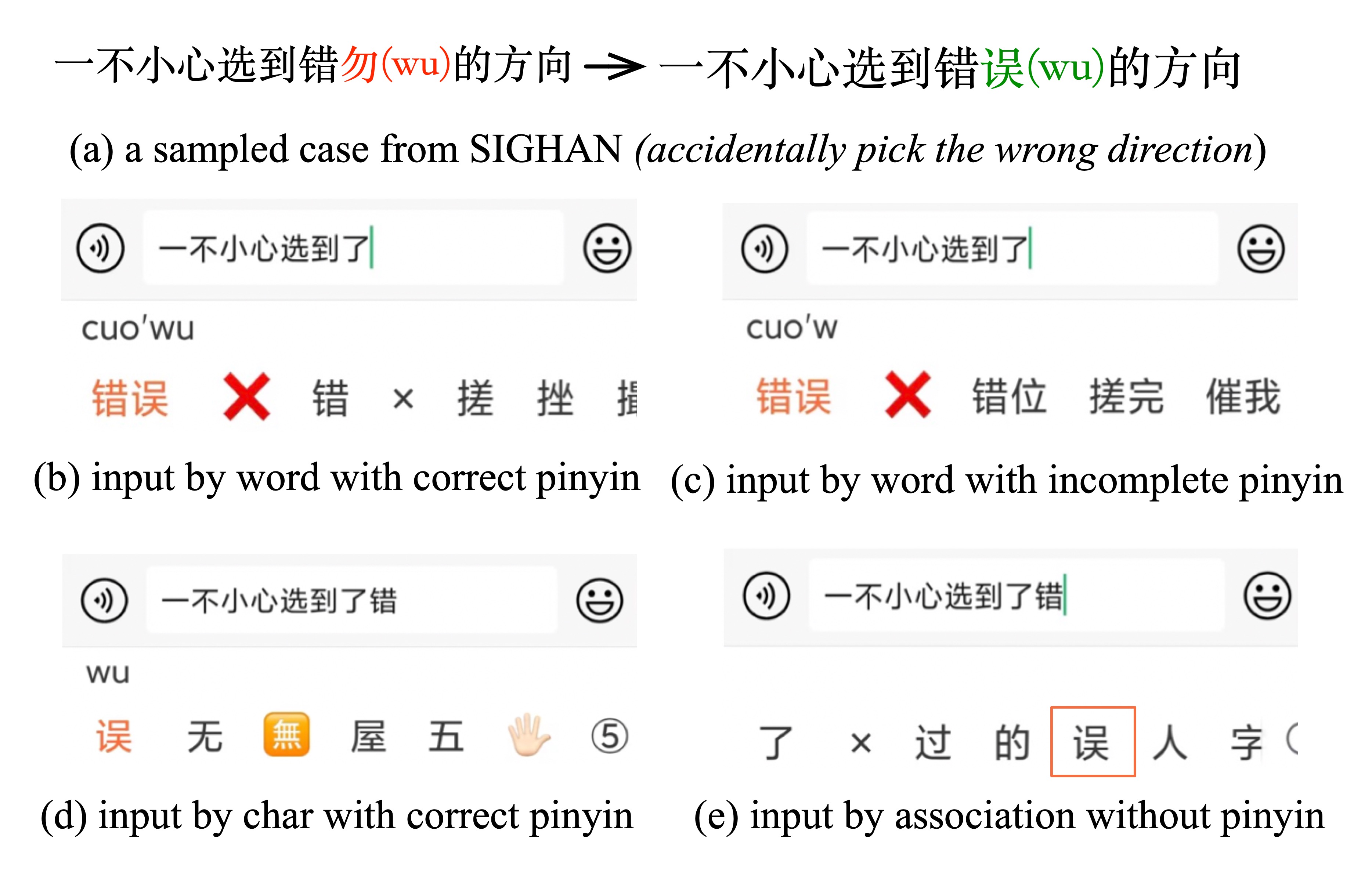}
  \caption{An error from SIGHAN: misspelling “错误” as “错勿”. Despite having the same pronunciation,  it's hard to reproduce this error in the given context through a Chinese IME, no matter what input form is used.}
  \label{Fig:problem}
\end{figure}
\end{CJK*}

However, there is still no CSC dataset specifically designed for native speakers. Existing CSC datasets, such as SIGHAN13, 14, and 15 \citep{ShihHungWu2013ChineseSC, LiangChihYu2014OverviewOS, YuenHsienTseng2015IntroductionTS}, are all sourced from Chinese learners. 
Spelling errors made by Chinese learners differ greatly from those made by native speakers. This is because Chinese input relies on Chinese input methods (IME), and modern Chinese IMEs always have powerful language models, making it difficult to recommend candidates that clearly do not fit the context. As shown in Figure \ref{Fig:problem}, native speakers using Chinese IMEs are unlikely to make such an unusual error.

Furthermore, the size of existing datasets is limited. As shown in Table \ref{Tab:statistic}, for three SIGHAN datasets, the training set contains an average of merely 2158 samples, while the test set comprises an average of only 1054 samples, and no development set is provided. When using such small-scale datasets, it is difficult for models to be trained sufficiently and for evaluation results to be reliable.

To address the aforementioned issues, we introduce CSCD-NS, a Chinese spelling check dataset designed for native speakers. The dataset is sourced from real Weibo (a Chinese social media platform) posts, which contain genuine spelling errors made by native speakers during their input process. Moreover, the dataset comprises 40,000 samples, which is ten times larger than previous datasets and this is also the largest dataset for the CSC task. To conduct an in-depth investigation into the distribution of spelling errors, we develop a tagging system that operates at phonetic and semantic levels. The analysis indicates that native speakers make a higher proportion of homophonic and word-level errors compared to Chinese learners, with the proportion of word-level errors doubling.

\begin{CJK*}{UTF8}{gbsn}
  \begin{figure}[t!]
    \centering
    \includegraphics[width=0.45\textwidth]{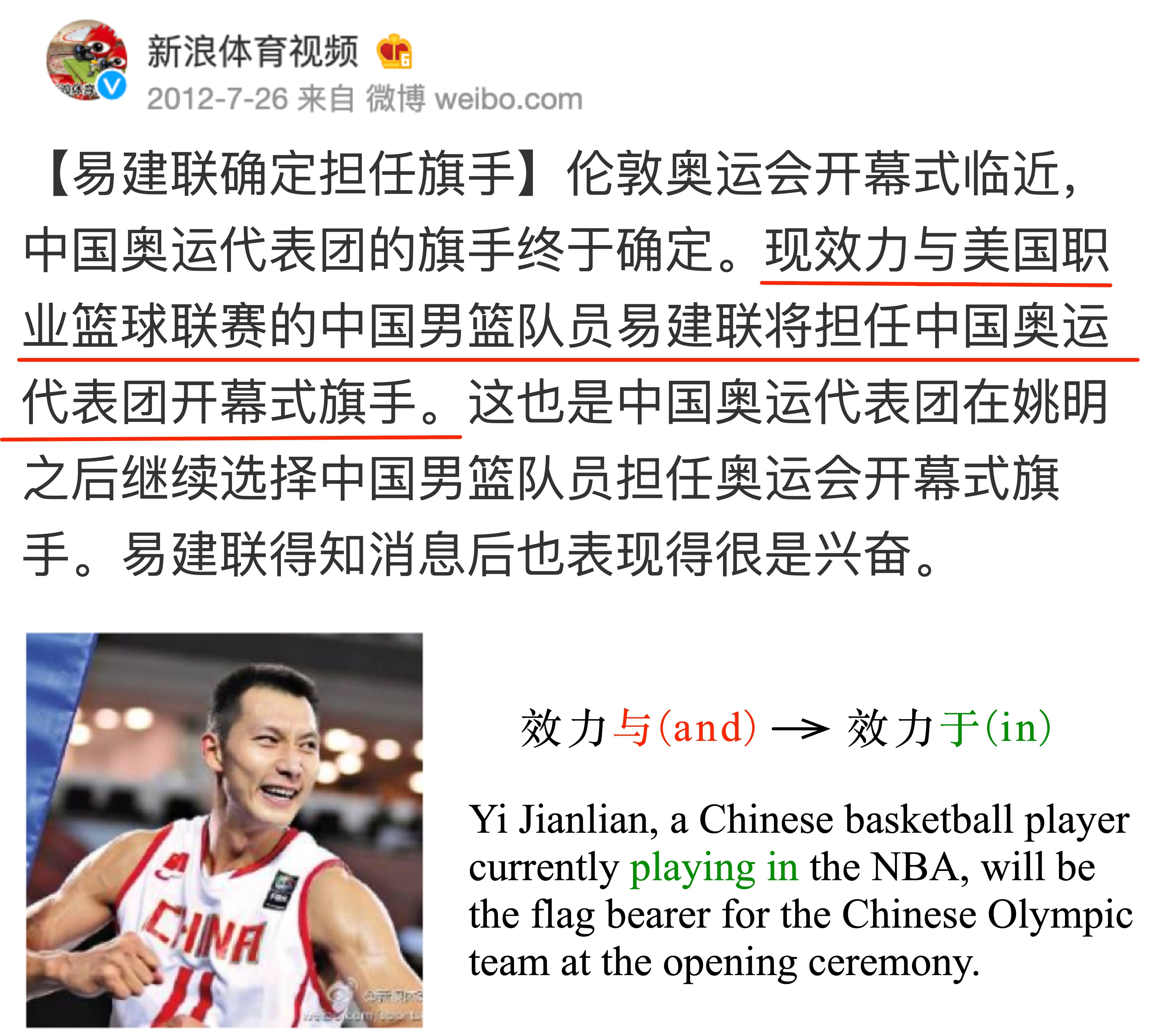}
    \caption{An authentic Weibo post from LCSTS, where the phrase "效力于" is mistakenly written as "效力与".}
    \label{Fig:example}
  \end{figure}
\end{CJK*}

Due to the lack of labeled data, previous studies always build additional pseudo data to improve the performance of models. However, these methods, which rely on confusion sets \citep{liu2021plome,zhang2020spelling} or ASR transcriptions \citep{wang2018hybrid}, do not align with the real-world input scenario. Therefore, we propose a novel method that directly simulates the input process through the Chinese IME and adds sampled noises to construct high-quality pseudo data. Experimental results show that our method can better fit the real error distribution and bring greater improvements.

We conduct comprehensive experiments on CSCD-NS, with different model sizes (from 0.1B to 13B parameters), architectures (encoder-only, encoder-decoder, and decoder-only), and learning approaches (fine-tuning and in-context learning). We also evaluate the performance of ChatGPT and GPT4.  The results demonstrate that BERT-like classification models outperform generative models, as the latter struggle with the simultaneous constraints of text length and pronunciation. Concurrently, the CSC task for native speakers is challenging due to the high proportion of word-level errors, leaving substantial room for improvement. 

In summary, our contributions are as follows:
\begin{itemize}
    \item We introduce the first Chinese spelling check dataset for native speakers which is also the largest dataset for the CSC task. Through quantitative analyses, we further unveil the specific error distribution for this scenario.
    \item We propose a novel method for constructing high-quality and large-scale pseudo data through a Chinese IME. Experimental results show that our method can bring greater improvements than existing methods.
    \item We explore the performance of different types of models in this scenario and analyze the challenges. To the best of our knowledge, we are the first to investigate the effectiveness and limitations of large language models (LLMs), such as ChatGPT, in addressing the CSC task.
\end{itemize}

\section{Related Work}

\textbf{CSC Datasets}: 
The existing CSC datasets, such as the SIGHAN series \citep{ShihHungWu2013ChineseSC, LiangChihYu2014OverviewOS, YuenHsienTseng2015IntroductionTS}, primarily cater to Chinese learners. However, these datasets suffer from limited data size and significant discrepancies in spelling errors compared to those made by native speakers. While there have been some efforts to develop Chinese grammatical error correction (CGEC) datasets for native speakers \citep{ma-etal-2022-linguistic, xu-etal-2022-fcgec, zhao2022overview, wang-etal-2022-cctc}, no such work has been undertaken for CSC datasets.

\textbf{CSC Data Augmentation}: In order to compensate for the lack of labeled data, previous studies often create additional pseudo data to enhance performance. The mainstream method is based on confusion sets \citep{liu2021plome, zhang2020spelling}, the pseudo data generated in this way is large in size but low in quality because context information is not considered.
Another relatively high-quality construction method is based on ASR \citep{wang2018hybrid}. However, this approach requires additional labeled ASR data, making it difficult to create large-scale datasets. Moreover, the spelling errors generated by these two methods differ greatly from those produced by native speakers, such as having a much smaller proportion of word-level errors. We provide a detailed analysis in Appendix A.

\textbf{CSC models}: In recent years, BERT-like \citep{devlin-etal-2019-bert} classification models have dominated the research of the CSC task 
\citep{hong2019faspell, zhu2022mdcspell, huang2021phmospell,zhang2020spelling, liu2021plome, liu2022craspell}. 
However, due to the lack of large-scale and high-quality datasets, the performance of these models is greatly limited.

\section{CSCD-NS}
In this section, we will show how to build CSCD-NS and discover the error distribution.

\subsection{Data Source}
We chose the LCSTS dataset \citep{hu-etal-2015-lcsts} as our data source. This dataset is composed of authentic Weibo posts, which is a popular Chinese social media platform. As shown in Figure \ref{Fig:example}, spelling errors found within these posts reflect the genuine mistakes made by native speakers during the input process. Furthermore, this dataset contains over 2 million posts and covers a wide range of fields, such as finance, sports, and entertainment. The substantial scale and scope of the LCSTS make it suitable to serve as the data source.

\subsection{Data Selection}
We split posts in LCSTS into sentences and obtain over 8 million sentences.
It is not realistic to label all of these sentences, and most of them are completely correct.
Therefore, we use an error detection model to filter out these correct sentences.

\textbf{Detection Model}: 
Given a source sequence ${\rm \bf X} = \{ x_{1}, x_{2}, ..., x_{N} \}$, the detection model is to check whether a token $x_{i}(1 \leq  i \leq N)$ is correct or not. 
We use the label 1 and 0 to mark the misspelled and the correct, respectively.
The detection model can be formalized as follows:
\begin{equation}
  {\rm \bf y} = sigmoid(W^{T}(E({\rm \bf e})))
\end{equation}
where ${\rm \bf e} = \{ e_{1}, e_{2}, ..., e_{N} \}$ is the sequence of word embeddings and ${E(*)}$ is the pre-trained encoder. 
The output ${\rm \bf y} = \{ y_{1}, y_{2}, ..., y_{N} \}$ is the sequence of probabilities, where $y_{{i}} \in (0,1)$ denotes the probability that $x_{i}$ is erroneous. 

\textbf{Training}: We follow the successful experience \citep{wang2020combining} of the NLPTEA2020 task \citep{rao2020overview} 
and use a Chinese ELECTRA-Large discriminator model \footnote{https://github.com/ymcui/Chinese-ELECTRA} \citep{clark2020electra} to initialize the detection model. 
Following previous research, we train the detection model on SIGHAN13-15's training data and Wang's pseudo data \citep{wang2018hybrid}
and save the best checkpoint by SIGHAN13-15's test data \footnote{SIGHAN datasets have no development set.}.

\textbf{Filtering}: We then use the trained detection model to filter out correct sentences. 
For the input sentence, we can obtain the error probability of each token ${\rm \bf y} = \{ y_{1}, y_{2}, ..., y_{N} \}$.
\begin{CJK*}{UTF8}{gbsn}
Previous research indicates that the detection model struggles with certain Chinese particles (的/地/得) due to the poor labeling of these words in SIGHAN datasets. Additionally, low-frequency entity words, such as person names, are also prone to over-checking. To address these issues, we utilize a Chinese lexical analysis tool (LAC) \citep{jiao2018LAC} to identify these particles and entities in the input sentence. We categorize tokens into three groups: 
$C_{particle}, C_{entity}, C_{others}$. Then, we calculate the maximum error probability for tokens in each category. If a category is empty, the maximum error probability is 0. We only consider a sentence correct if all the maximum error probabilities for each category are below the corresponding threshold. This can be formalized as follows:
\begin{equation}
  \left\{
\begin{aligned}
  & max(\{y_{i} | x_{i} \in C_{particle}\}) < \delta_{particle} \\
  & max(\{y_i | x_{i} \in C_{entity}\}) < \delta_{entity} \\
  & max(\{y_i | x_{i} \in C_{others}\}) < \delta_{others}
\end{aligned}
\right.
\end{equation}
Here, $\delta_{particle}$, $\delta_{entity}$ and $\delta_{others}$ represent threshold values. 
These thresholds are determined using a small manually labeled set and are set to 0.05, 0.5, and 0.15 respectively.
\end{CJK*} 

Based on the above method, we filter out approximately 91.2\% of sentences, retaining around 700,000 sentences that may contain spelling errors. To verify the accuracy of our filtering, we randomly select 2,000 filtered sentences and find that the accuracy is 99.2\%, aligning with our expectations. For the remaining sentences, we randomly select a portion for manual annotation.

\begin{table*}
\small 
\centering
\def\arraystretch{1.3}
\begin{tabular}{ccccccccc}
\hline
Dataset & Train Size & Dev Size & Test Size & Target Group & Source & Language &  Err. ratio & Avg err./sent. \\
\hline
SIGHAN13 & 700 & - & 1000 & Chinese learners & essays & TC & 77.11\% & 1.20 \\
SIGHAN14 & 3437 & - & 1062 & Chinese learners & essays & TC & 86.19\% & 1.52 \\
SIGHAN15 & 2339 & - & 1100 & Chinese learners & essays & TC & 81.82\% & 1.33 \\
\hline
CSCD-NS & 3,0000 & 5,000 & 5,000 & native speakers & tweets & CN & 46.02\% & 1.09 \\
\hline
\end{tabular}
\caption{The comparison of CSCD-NS and existing CSC datasets SIGHAN13, SIGHAN14, and SIGHAN15 in terms of dataset size, target group, data source, language, error sentence ratio, and average errors per sentence. In the table, TC and CN respectively denote Traditional Chinese and Simplified Chinese.}
\label{Tab:statistic}
\end{table*}

\begin{CJK*}{UTF8}{gbsn}
\begin{table*}
\small 
\centering
\def\arraystretch{1.3}
\begin{tabular}{ccccccc}
\hline
origin & \multicolumn{6}{c}{由\textcolor{red}{之}可见，中国企业的技术提升后，\textcolor{red}{因}与跨国企业共同研发，不\textcolor{red}{在}简单的代加工} \\
correct & \multicolumn{6}{c}{由\textcolor{blue}{此}可见，中国企业的技术提升后，\textcolor{blue}{应}与跨国企业共同研发，不\textcolor{blue}{再}简单的代加工} \\
segment & \multicolumn{6}{c}{由此可见\; ，\;中国\;企业\;的\;技术\;提升\;后\;，\; 应\;与\;跨国企业\;共同研发\;，\;不再\;简单\;的\;代加工} \\
translation & \multicolumn{6}{c}{\tabincell{l}{It can be seen that after the technology of Chinese enterprises is upgraded, \\they should cooperate with multinational enterprises in research instead of simple processing. }}\\
\hline
\multirow{4}{*}{errors} & word pair & pinyin pair (ed) & phonetic tag & word len & ori-word valid & semantic tag   \\
\cline{2-7}
& 由\textcolor{red}{之}可见 $\rightarrow$ 由\textcolor{blue}{此}可见 & zhi $\rightarrow$ ci (2) & dissimilar & 4 & \XSolidBrush & character \\
& \textcolor{red}{因} $\rightarrow$ \textcolor{blue}{应} & yin $\rightarrow$ ying (1) & similar & 1 &  - & character \\
& 不\textcolor{red}{在} $\rightarrow$ 不\textcolor{blue}{再} & zai $\rightarrow$ zai (0) & same & 2 & \Checkmark & word \\
\hline
\end{tabular}
\caption{The process of adding phonetic and semantic tags. In the table, "ed" means edit distance, and "ori-word valid" indicates the validity of the original word.}
\label{Tab:tag_system}
\end{table*}
\end{CJK*}

\subsection{Data Annotation}
We recruit a group of native speakers for manual annotation. The annotators are required to check whether the given sentence contains any spelling errors and provide the correct sentence. To ensure the quality of annotation, each sentence is annotated at least twice by different annotators. If the results of the two annotations are inconsistent, a senior annotator will make the final decision.

To clarify the annotation rules and reduce disputes during the annotation process, sentences that fall into the following three categories will be directly discarded:
(1) sentences with inherent ambiguity;
(2) sentences with multiple reasonable answers to errors;
(3) sentences with complex grammatical errors.
Therefore, the sentence retained in the annotation process is semantically clear and has a unique correction result.

In the end, we obtain 40,000 manually annotated sentences, which constitute the CSCD-NS dataset. After random partitioning, there are 30,000 samples in the training set, and 5,000 samples each in the development and test sets.

\subsection{Analysis on Basic Statistics}
\label{section:statistic}
As shown in Table \ref{Tab:statistic}, the CSCD-NS is significantly larger in scale compared to existing datasets. Moreover, only the CSCD-NS provides a development set, is in Simplified Chinese, and originates from daily input by native speakers. Additionally, the CSCD-NS exhibits a more balanced distribution of positive and negative samples, with fewer spelling errors per sentence on average, suggesting a lower error rate among native speakers compared to Chinese learners.

\subsection{Analysis on Error Distribution}
To conduct an in-depth study on the differences between native speakers and Chinese learners in terms of spelling errors, we design a tagging system for quantitative analyses.

\textbf{Tag definition}: We define three phonetic-level tags and two semantic-level tags. 
The phonetic tags consist of:
(1) same phonetic error: the erroneous character has the same pronunciation as the correct one.
(2) similar phonetic error: the erroneous character's pronunciation has an edit distance of 1 from the correct character's pronunciation.
(3) dissimilar phonetic error: the erroneous character's pronunciation has an edit distance greater than 1 from the correct character's pronunciation. The semantic tags consist of:
(1) word-level error: the erroneous word is a valid Chinese word.
(2) character-level error: the erroneous word is not a valid Chinese word, or the length of the erroneous word is 1.

As shown in Table \ref{Tab:tag_system}, we first tokenize the correct sentence using LAC \citep{jiao2018LAC} to obtain word-level correction pairs. For each pair, we compute the pinyin edit distance and assign a phonetic-level tag. Simultaneously, we check the original word's validity in Chinese and incorporate its length to assign a semantic tag.

\begin{figure*}[htbp]
	\centering
	\includegraphics[width=0.95\textwidth]{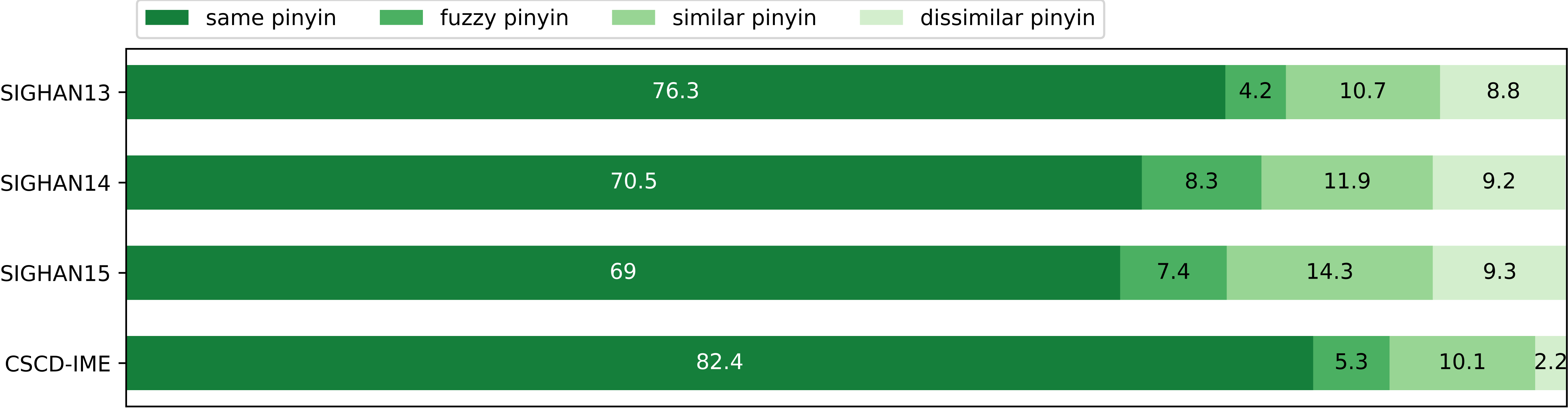}
	\includegraphics[width=0.95\textwidth]{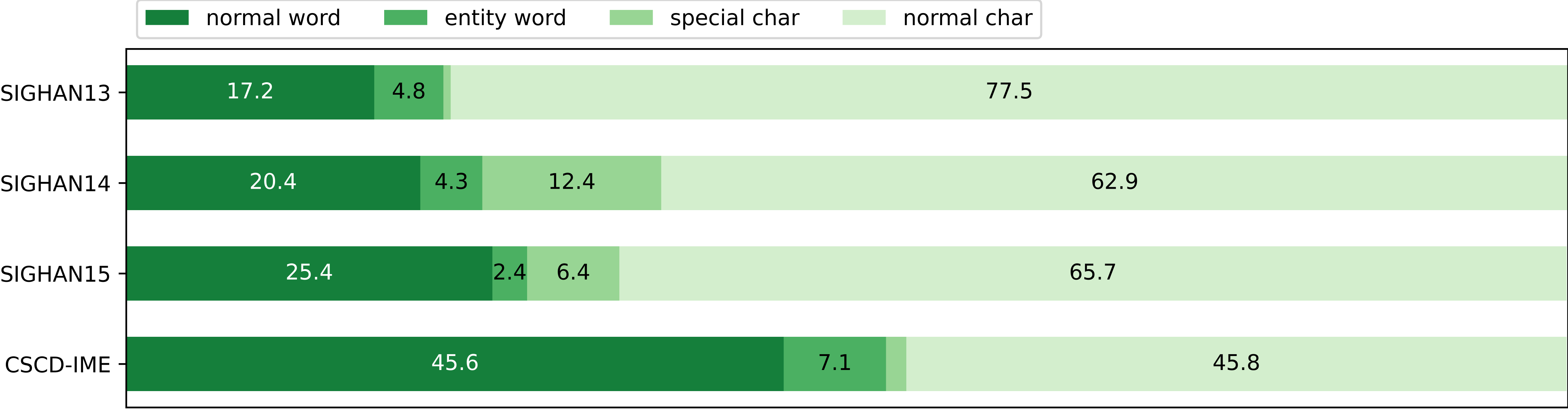}
	\caption{The comparison of error distribution (\%) at phonetic level (above) and semantic level (below).}
	\label{fig:analysis}
\end{figure*}

\textbf{Phonetic-level analysis}: As illustrated in Figure \ref{fig:analysis}, the proportion of same phonetic errors is the largest, while the proportion of dissimilar phonetic errors is the smallest in all four datasets. This feature is more pronounced in the CSCD-NS dataset, where the proportion of dissimilar phonetic errors is only 2.2\%, significantly lower than in the other datasets. Over 97\% of the errors are either the same phonetic or similar phonetic errors. This is because even if users make slight mistakes in their pinyin input, Chinese IME will auto-fix the input pinyin based on the context \citep{jia2014joint}.

\textbf{Semantic-level analysis}: As shown in Figure \ref{fig:analysis}, the proportion of word-level errors in CSCD-NS (49.4\%) far exceeds that of existing datasets, which is twice the average value (23.3\%) of the SIGHAN datasets. This is because native speakers rely on the IME to input Chinese texts, which tends to recommend relatively reasonable valid words rather than strange "error words", resulting in a lower proportion of character-level errors. Compared to character-level errors, word-level errors pose a greater challenge to CSC systems.

\section{Data Augmentation}
The manual annotation of CSC dataset is very expensive, therefore, how to construct pseudo data has always been a valuable topic.
In this section, we introduce a novel method that can generate high-quality pseudo data on a large scale.

\subsection{Data Preparation}
The basic principle of pseudo-data construction is to add noise to accurate sentences.
Therefore, it is necessary to first prepare completely correct sentences. Fortunately, such text data is readily available on the Internet, including Wikipedia articles and classic books. This availability also ensures the generation of a large-scale dataset.

\begin{figure*}[htbp]
	\centering
	\includegraphics[width=0.95\textwidth]{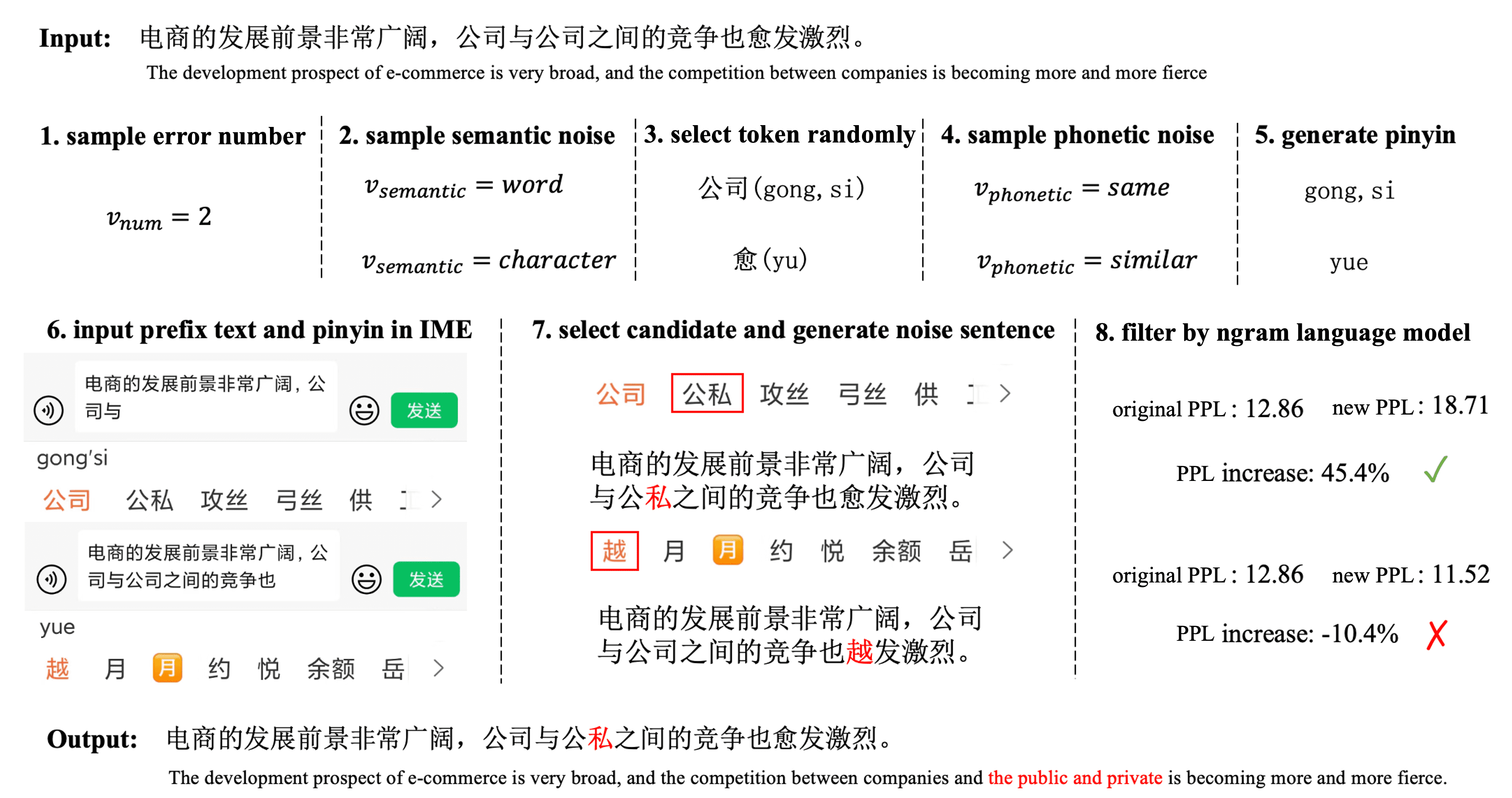}
	\caption{The IME-based pseudo data generation process.}
	\label{fig:pseduo}
\end{figure*}

\subsection{IME-based Pseudo Data Generation}
First, we should analyze and obtain the error distribution based on the annotated data, including the distribution of the number of errors per sentence $D_{num}$, phonetic-level error distribution $D_{phonetic}$, and semantic-level error distribution $D_{semantic}$. 

As illustrated in Figure \ref{fig:pseduo}, the IME-based generation of pseudo data involves eight steps. 

(1) Sample a noise $v_{num}$ based on $D_{num}$, which indicates the number of generated spelling errors. The following steps are performed for each error.

(2) Sample a semantic noise $v_{semantic}$ based on $D_{semantic}$, which indicates whether the error is at the word level or the character level.

(3) Randomly select a token from the original text based on the sampled $v_{semantic}$.

(4) Sample a phonetic noise $v_{phonetic}$ based on $D_{phonetic}$, which indicates whether the error is the same, similar, or dissimilar phonetic error.

(5) Generate the new pinyin $p$, based on the sampled phonetic noise $v_{phonetic}$ and the actual pronunciation of the selected token.

(6)  In a Chinese IME, input the correct text before the selected token $t$ and enter the generated pinyin $p$. The IME would then recommend reasonable candidates $\{c_1, c_2, ..., c_n\}$. Leveraging the powerful language model of the IME, candidates are recommended by considering both the context before token $C_{<t}$ and the pronunciation $p$ \citep{chen2015neural}. This can be represented as: 

\begin{equation}
    \{c_1, c_2, ..., c_n\} = {\rm IME}(C_{<t}, p)
\end{equation}

(7) Choose the candidate from the recommendations. If the first recommended candidate is the original token, randomly select the second or third candidate word $\{c_2, c_3\}$. If the first candidate word is not the original token, directly choose the first candidate word $c_1$. Then, replace the original token in the input text with the selected candidate word to generate a noisy sentence.

(8) Due to the powerful language model of IME, the generated sentence may still be a correct sentence. Therefore, we adopt an n-gram language model for secondary filtering. We consider the generated sentence to be incorrect only if its perplexity (PPL) exceeds that of the original sentence by a threshold of $\delta$. This can be formalized as follows:
    
\begin{equation}
    \frac{PPL(noisy) - PPL(origin)}{PPL(origin)} > \delta
\end{equation}
    
Through these steps, we can generate pseudo data that closely resembles the actual input process.

\subsection{LCSTS-IME-2M}
We apply the above method to construct a large-scale CSC pseudo dataset LCSTS-IME-2M, consisting of about 2 million samples, based on the correct sentences filtered from LCSTS, the error distribution of CSCD-NS, and the Google IME \footnote{https://www.google.com/inputtools/}.

\section{Experiments}
In this section, we evaluate the performance of different models on CSCD-NS and compare different pseudo-data construction methods.

\begin{table}
\small 
\centering
\def\arraystretch{1.3}
\begin{tabular}{lccc}
\hline
Model & Structure & Parameters & Learning \\
\hline
BERT & Encoder & 102M & FT \\
SM BERT & Encoder & 123M & FT \\
PLOME & Encoder & 123M & FT \\
BART & En-Decoder & 407M & FT \\
Baichuan2-7B & Decoder & 7.5B & LoRA \\
Baichuan2-13B & Decoder & 13.9B & LoRA \\
ChatGPT & Decoder & - & ICL \\
GPT4 & Decoder & - & ICL \\
\hline
\end{tabular}
\caption{The comparison of different baselines. In the table, En-Decoder refers to encoder-decoder, FT refers to full-parameter finetuning, LoRA refers to finetuning using low-rank adaptation, and ICL refers to in-context learning. Note that the number of parameters for ChatGPT and GPT4 has not been disclosed by the official documentation.}
\label{Tab:model}
\end{table}

\begin{table*}
  \small 
  \centering
  \bgroup
  \setlength\tabcolsep{0.6em}
  \def\arraystretch{1.2}
  \begin{tabular}{l|ccc|ccc|ccc|ccc}
  \hline
  \multirow{3}{*}{Models} & \multicolumn{6}{c|}{Sentence level} & \multicolumn{6}{c}{Character level} \\
  \cline{2-13}
  & \multicolumn{3}{|c|}{Detection} & \multicolumn{3}{|c|}{Correction} & \multicolumn{3}{|c|}{Detection} & \multicolumn{3}{|c}{Correction} \\
  \cline{2-13}
  & P & R & F1 & P & R & F1 & P & R & F1 & P & R & F1 \\
  \hline
  BERT & 79.16 & 65.83 & 71.88 & 70.55 & 58.66 & 64.06 & 83.00 & 67.01 & 74.15 & 73.59 & 59.41 & 65.75 \\ 
  \footnotesize +\textit{LCSTS-IME-2M} & 78.98 & 73.60 & 76.20 & 75.63 & 70.47 & 72.96 & 82.19 & 75.75 & 78.84 & 78.84 & 72.67 & 75.63 \\
  \hline
  {SM BERT} & 80.87 & 64.78 & 71.94 & 74.42 & 59.62 & 66.20 & \textbf{84.46} & 65.35 & 73.68 & 77.50 & 59.97 & 67.62 \\
  {\footnotesize +\textit{LCSTS-IME-2M}} & 79.19 & \textbf{74.86} & \textbf{76.97} & 75.75 & \textbf{71.60} & 73.62 & 82.39 & \textbf{77.93} & \textbf{80.10} & 78.63 & \textbf{74.37} & 76.44 \\
  \hline
  {PLOME} & 79.78 & 57.23 & 66.65 & 78.09 & 56.01 & 65.23 & 83.48 & 57.99 & 68.44 & 81.49 & 56.61 & 66.81 \\ 
  {\footnotesize +\textit{LCSTS-IME-2M}} & \textbf{81.20} & 72.21 & 76.44 & \textbf{79.05} & 70.30 & \textbf{74.42} & 84.21 & 73.81 & 78.67 & \textbf{82.00} & 71.88 & \textbf{76.60} \\
  \hline
  BART & 38.73 & 46.05 & 42.08 & 35.41 & 42.11 & 38.47 & 36.97 & 63.32 & 46.69 & 33.30 & 57.04 & 42.05 \\
  {\footnotesize +\textit{LCSTS-IME-2M}} & 42.06 & 54.29 & 47.40 & 41.01 & 52.95 & 46.22 & 40.87 & 75.97 & 53.15 & 39.68 & 73.75 & 51.60 \\
  \hline
  Baichuan2-7B & 64.98 & 53.04 & 58.41 & 62.70 & 51.17 & 56.35 & 57.10 & 56.92 & 57.01 & 54.72 & 54.55 & 54.63 \\
  {\footnotesize +\textit{LCSTS-IME-2M}} & 66.94 & 66.13 & 66.54 & 64.84 & 64.05 & 64.44 & 60.63 & 72.57 & 66.07 & 58.55 & 70.08 & 63.80 \\
  \hline
  Baichaun2-13B & 67.53 & 60.23 & 63.67 & 65.14 & 58.11 & 61.42 & 60.07 & 64.62 & 62.26 & 57.49 & 61.86 & 59.60 \\
  {\footnotesize +\textit{LCSTS-IME-2M}} & 67.82 & 67.35 & 67.58 & 66.33 & 65.87 & 66.10 & 61.67 & 73.91 & 67.24 & 60.06 & 71.98 & 65.48 \\
  \hline
  ChatGPT & 59.74 & 51.60 & 55.38 & 55.17 & 47.66 & 51.14 & 60.41 & 55.73 & 57.98 & 54.84 & 50.59 & 52.63 \\
  \hline
  GPT4 & 58.37 & 59.71 & 59.03 & 53.67 & 54.90 & 54.28 & 58.40 & 63.60 & 60.89 & 52.34 & 57.00 & 54.57 \\
  \hline
\end{tabular}
\egroup
\caption{The performance (\%) of different models on CSCD-NS with or without pseudo dataset.}
\label{Tab:result}
\end{table*}

\begin{table}
  \small 
  \centering
  \def\arraystretch{1.2}
  \begin{tabular}{cccc}
  \hline
  Models & Char level & Word level & $\Delta$ \\
  \hline
  BERT & 72.82 & 71.07 & -1.75 \\
  SM BERT & 75.09 & 72.71 & -2.38 \\
  PLOME & 77.77 & 72.78 & -4.99 \\
  BART & 57.19 & 55.60 & -1.59 \\
  Baichaun2-7B & 65.88 & 63.50 & -2.38 \\
  Baichaun2-13B & 71.58 & 68.88 & -2.70 \\
  ChatGPT & 61.96 & 57.65 & -4.31 \\
  GPT4 & 71.06 & 61.13 & \textbf{-9.93} \\ 
  \hline
\end{tabular}
  \caption{The performance (correction F1 score at character level \%) comparison between word-level and character-level errors. We only select the same phonetic errors here to avoid the influence of pronunciation.}
\label{Tab:word-level}
\end{table}

\begin{CJK*}{UTF8}{gbsn}
\begin{table*}
\small 
\centering
\def\arraystretch{1.3}
\begin{tabular}{cl}
\hline
origin &  新方案还处多方博弈中，想要尽快\textcolor{red}{的}打破僵局仍\textcolor{red}{就}困难重重，我们会跟紧并持续报\textcolor{red}{到} \\
correct & 新方案还处多方博弈中，想要尽快\textcolor{blue}{地}打破僵局仍\textcolor{blue}{旧}困难重重，我们会跟紧并持续报\textcolor{blue}{道} \\
translation & \tabincell{l}{The new plan is still in a multi-party game. It is still difficult to break the deadlock as soon as possible. \\ We will follow up and continue to report.}\\
\hline
PLOME & 仍\textcolor{red}{就}(jiu) $\rightarrow$ 仍\textcolor{blue}{旧}(jiu); \ 跟\textcolor{red}{紧}(jin) $\rightarrow$ 跟\textcolor{blue}{进}(jin)  \\
\hline
ChatGPT & 处 $\rightarrow$ 处\textcolor{blue}{于}; \  尽快\textcolor{red}{的}(de) $\rightarrow$ 尽快\textcolor{blue}{地}(de) ; \ 仍\textcolor{red}{就}(jiu) $\rightarrow$ 仍\textcolor{blue}{然}(ran); \ 跟\textcolor{red}{紧}(jin) $\rightarrow$ 跟\textcolor{blue}{进}(jin) \\
\hline
\end{tabular}
\caption{The correction results of PLOME and ChatGPT. The pronunciation of the character is in brackets.}
\label{Tab:error_analysis}
\end{table*}
\end{CJK*}

\begin{table}
  \small 
  \centering
  \def\arraystretch{1.2}
  \begin{tabular}{ccccc}
  \hline
  Data & BERT & SM BERT & BART & Baichuan2-7B \\
  \hline
  *CS & 19.57 & 15.39 & 14.02 & 25.67 \\
  *ASR & 42.22 & 39.50 & 29.97 & 35.69 \\
  *IME & \textbf{46.71} & \textbf{53.84} & \textbf{32.16} & \textbf{38.64}\\
  \hline
  +CS & 64.53 & 67.36 & 42.95  & 54.30 \\
  +ASR & 68.44 & 71.26 & 44.88 & 56.77\\
  +IME & \textbf{70.41} & \textbf{72.72} & \textbf{45.92} & \textbf{57.85}\\
  \hline
\end{tabular}
  \caption{The comparison of the performance (correction F1 score at character level \%) of three pseudo-data construction methods based on confusion sets (CS), ASR, and IME. In the table, an asterisk (*) indicates that only pseudo data is used for training, while a plus sign (+) denotes pretraining on pseudo data followed by continued training on the CSCD-NS's training data.}
\label{Tab:pseudo}
\end{table}

\subsection{Basic Settings}
\textbf{Data}:
We perform experiments based on the labled data CSCD-NS and the pseudo data LCSTS-IME-2M. For pseudo data, we pre-train the model on it first, then fine-tune the model on the labeled data.

\textbf{Metric}:
We compute detection and correction metrics at the sentence level and character level, including precision, recall, and F1 score. For sentence-level metrics, we use the calculation method in FASPell \citep{hong2019faspell}. For character-level metrics, we calculate all characters instead of only those correctly detected characters.

\textbf{Baselines}: As shown in Table \ref{Tab:model}, the baselines encompass a diverse range of model structures, sizes, and learning methods. (1) BERT \citep{devlin-etal-2019-bert} directly fine-tunes the standard masked language model to generate fixed-length corrections. (2) Soft-Masked BERT (SM BERT) \citep{zhang2020spelling} employs an error detection model to provide better correction guidance. (3) PLOME \citep{liu2021plome} integrates phonetic and visual features into the pre-trained model. It has included a pre-training step on a confusion set-based pseudo dataset. (4) BART \citep{lewis-etal-2020-bart} models the CSC as a sequence-to-sequence task. We use the Chinese BART-large version here \footnote{https://huggingface.co/fnlp/bart-large-chinese}. (5) Baichuan2 \citep{baichuan2023baichuan2} models the CSC as a text generation task based on instructions. We fine-tune the model by LoRA \citep{hu2021lora} and use the version of 7B and 13B here \footnote{https://github.com/baichuan-inc/Baichuan2}. (6) ChatGPT and GPT4 perform the CSC task in a few-shot setting (10 examples) through in-context learning (ICL) \citep{dong2022survey}.

To ensure that the correction results are of the same length as the input text, we only extract equal-length substitution modifications for generative models (BART, Baichuan2, ChatGPT and GPT4). Further implementation details of these models can be found in Appendix B.

\subsection{Main Results}
(1) As shown in Table \ref{Tab:result}, compared with generative models, BERT-like token-level classification models (BERT, SM BERT, PLOME) remain the best approach for the CSC task, with smaller model size, higher performance, and faster inference speed.

(2) The overall performance of generative models is relatively poor because the CSC task has strong constraints, requiring corrections to be of equal length and phonetically similar to the original text. These strong constraints make it easy for generative models to cause over-correction and incorrect correction. 

(3) For generative models, as the parameter size increases, their performance tends to improve gradually. 
This trend can be observed from smaller models like BART (0.4B) to larger ones such as Baichuan2-13B. Similarly, GPT4 outperforms ChatGPT, and it is only through in-context learning that GPT4 can achieve performance comparable to Baichuan2-7B fine-tuned on CSCD-NS.

(4) Large-scale and high-quality pseudo data is important for improving the performance, bringing consistent improvements across all six models.

(5) The task of CSC for native speakers is highly challenging and the best F1 score of baseline models is still below 80. A key characteristic of this scenario is the high proportion of word-level errors. As shown in Table \ref{Tab:word-level}, word-level errors are more difficult for models to handle than character-level errors, as they require understanding more complex contexts. The development of CSC models, from BERT to PLOME, has primarily focused on optimizing character-level errors, with little progress made in addressing word-level errors. Therefore, further efforts are required in this scenario.

\subsection{Better Data Augmentation Method}
In this part, we compare different pseudo-data construction methods. We conduct experiments on an existing ASR-based pseudo dataset \citep{wang2018hybrid}, containing about 271K samples. We extract the correct sentences and construct new pseudo-data based on confusion sets and IME, respectively.

As demonstrated in Table \ref{Tab:pseudo}, our IME-based approach exhibits a substantial enhancement in performance compared to the other two methods. This improvement is even more pronounced when training exclusively on pseudo-data. The primary factor contributing to this success is the error distribution. As depicted in Figure \ref{fig:analysis-pseudo}, the pseudo-data generated via the IME-based method more accurately reflects the spelling errors made by native speakers. More analysis can be found in Appendix A.

\subsection{Discussions}
\begin{CJK*}{UTF8}{gbsn}
For generative models, it is difficult to ensure that the generated text satisfies constraints on length and pronunciation. In the original correction results produced by ChatGPT, a staggering 82.1\% of modifications exhibit unequal length, while 35.4\% display dissimilar pronunciation. As illustrated in Table \ref{Tab:error_analysis}, the replacement of "处" with "处于" (located in) disregards the length constraint by introducing an additional character. Similarly, the correction of "仍旧" to "仍然" (still) overlooks the pronunciation constraint. Although these alterations may appear reasonable, they fail to meet the CSC task's requirements.

BERT-like classification models have difficulty in addressing complex word-level errors and equal-length grammatical errors, as these require a strong contextual understanding. For example, the PLOME model shows a recall rate of only 60\% for word-level errors and merely 44\% for particle-related grammatical errors (的/地/得). Table \ref{Tab:error_analysis} illustrates that the incorrect word "报到" (check-in) is a high-frequency term, necessitating the model to recognize its context and correct it to "报道" (report). Similarly, in the phrase "尽快的打破" (try to break), the model must comprehend the grammatical rule (the particle between the adjective and the verb should be "地" instead of "的") and apply the appropriate correction.

Moreover, all baseline systems, which are based on pre-trained language models, exhibit a propensity to over-convert low-frequency expressions into more prevalent ones \citep{zhang2020spelling, liu2022craspell}. As demonstrated in Table \ref{Tab:error_analysis}, "跟紧" and "跟进" share similar meanings (follow-up); however, since "跟进" is more frequently used, the model is prone to over-correcting.

Consequently, enabling controlled text generation, addressing complex word-level and grammatical errors, and enhancing the understanding of low-frequency or new words all represent valuable avenues for future research.
\end{CJK*}

\section{Conclusion}
In this paper, we focus on CSC for native speakers. For this scenario, we propose a new dataset, CSCD-NS, which is also the largest dataset for CSC. We further unveil the specific error distribution, with a significantly higher proportion of word-level errors. 
Moreover, we introduce an IME-based pseudo-data construction approach, enabling large-scale generation of high-quality pseudo-data. We explore the performance of various models and first evaluate ChatGPT and GPT4 on the CSC task. Our experiments demonstrate that BERT-like models exhibit better performance than generative models, but there is still a considerable room for improvement. We hope these data resources and our findings could stimulate further research in this area.

\section{Limitations}
\begin{CJK*}{UTF8}{gbsn}
Limitation of the CSCD-NS dataset: The data source for the CSCD-NS dataset is derived from a Chinese social networking platform. Therefore, it may not fully represent the error distribution of native speakers, as there may be slight differences in other scenarios, such as formal document writing.

Limitation of the pseudo-data construction: The employed method of input simulation via IME is relatively basic, and the actual input scenario is more complex. For instance, individuals may utilize abbreviated pinyin to input common phrases, entering only the initials of characters (e.g., "wm" for "我们") \citep{tan2022exploring}. Moreover, a substantial number of users prefer the T9-style keyboard when employing IME on mobile devices. These factors collectively contribute to the inability of our pseudo-data construction method to accurately simulate the realistic input scenario.
\end{CJK*}

\section{Ethics Statement}
License: CSCD-NS and the constructed pseudo-data \textit{LCSTS-IME-2M} are based on LCSTS \citep{hu-etal-2015-lcsts}, we applied for and obtained the right to use this dataset, and performed the academic research under the copyright.

Annotator Compensation: In this work, annotators are from a data labeling company in China. Through the pre-labeling, we estimate that each annotator could label 80 samples per hour and the label speed would be faster when they are skilled. In China, 60 yuan (8.76 dollars) per hour is a fair wage, therefore, we pay the annotator 0.75 yuan (0.11 dollars) for each sentence.

\bibliography{anthology,custom}

\begin{thebibliography}{28}
\expandafter\ifx\csname natexlab\endcsname\relax\def\natexlab#1{#1}\fi

\bibitem[{Baichuan(2023)}]{baichuan2023baichuan2}
Baichuan. 2023.
\newblock \href {https://arxiv.org/abs/2309.10305} {Baichuan 2: Open
  large-scale language models}.
\newblock \emph{arXiv preprint arXiv:2309.10305}.

\bibitem[{Chen et~al.(2015)Chen, Zhao, and Wang}]{chen2015neural}
Shenyuan Chen, Hai Zhao, and Rui Wang. 2015.
\newblock Neural network language model for chinese pinyin input method engine.
\newblock In \emph{Proceedings of the 29th Pacific Asia conference on language,
  information and computation}, pages 455--461.

\bibitem[{Clark et~al.(2020)Clark, Luong, Le, and Manning}]{clark2020electra}
Kevin Clark, Minh-Thang Luong, Quoc~V. Le, and Christopher~D. Manning. 2020.
\newblock \href {https://openreview.net/pdf?id=r1xMH1BtvB} {{ELECTRA}:
  Pre-training text encoders as discriminators rather than generators}.
\newblock In \emph{ICLR}.

\bibitem[{Devlin et~al.(2019)Devlin, Chang, Lee, and
  Toutanova}]{devlin-etal-2019-bert}
Jacob Devlin, Ming-Wei Chang, Kenton Lee, and Kristina Toutanova. 2019.
\newblock \href {https://doi.org/10.18653/v1/N19-1423} {{BERT}: Pre-training of
  deep bidirectional transformers for language understanding}.
\newblock In \emph{Proceedings of the 2019 Conference of the North {A}merican
  Chapter of the Association for Computational Linguistics: Human Language
  Technologies, Volume 1 (Long and Short Papers)}, pages 4171--4186,
  Minneapolis, Minnesota. Association for Computational Linguistics.

\bibitem[{Dong et~al.(2022)Dong, Li, Dai, Zheng, Wu, Chang, Sun, Xu, and
  Sui}]{dong2022survey}
Qingxiu Dong, Lei Li, Damai Dai, Ce~Zheng, Zhiyong Wu, Baobao Chang, Xu~Sun,
  Jingjing Xu, and Zhifang Sui. 2022.
\newblock A survey for in-context learning.
\newblock \emph{arXiv preprint arXiv:2301.00234}.

\bibitem[{Hong et~al.(2019)Hong, Yu, He, Liu, and Liu}]{hong2019faspell}
Yuzhong Hong, Xianguo Yu, Neng He, Nan Liu, and Junhui Liu. 2019.
\newblock Faspell: A fast, adaptable, simple, powerful chinese spell checker
  based on dae-decoder paradigm.
\newblock In \emph{Proceedings of the 5th Workshop on Noisy User-generated Text
  (W-NUT 2019)}, pages 160--169.

\bibitem[{Hu et~al.(2015)Hu, Chen, and Zhu}]{hu-etal-2015-lcsts}
Baotian Hu, Qingcai Chen, and Fangze Zhu. 2015.
\newblock \href {https://doi.org/10.18653/v1/D15-1229} {{LCSTS}: A large scale
  {C}hinese short text summarization dataset}.
\newblock In \emph{Proceedings of the 2015 Conference on Empirical Methods in
  Natural Language Processing}, pages 1967--1972, Lisbon, Portugal. Association
  for Computational Linguistics.

\bibitem[{Hu et~al.(2021)Hu, Shen, Wallis, Allen-Zhu, Li, Wang, Wang, and
  Chen}]{hu2021lora}
Edward~J Hu, Yelong Shen, Phillip Wallis, Zeyuan Allen-Zhu, Yuanzhi Li, Shean
  Wang, Lu~Wang, and Weizhu Chen. 2021.
\newblock Lora: Low-rank adaptation of large language models.
\newblock \emph{arXiv preprint arXiv:2106.09685}.

\bibitem[{Huang et~al.(2021)Huang, Li, Jiang, Zhang, Chen, Wang, and
  Xiao}]{huang2021phmospell}
Li~Huang, Junjie Li, Weiwei Jiang, Zhiyu Zhang, Minchuan Chen, Shaojun Wang,
  and Jing Xiao. 2021.
\newblock Phmospell: Phonological and morphological knowledge guided chinese
  spelling check.
\newblock In \emph{Proceedings of the 59th Annual Meeting of the Association
  for Computational Linguistics and the 11th International Joint Conference on
  Natural Language Processing (Volume 1: Long Papers)}, pages 5958--5967.

\bibitem[{Jia and Zhao(2014)}]{jia2014joint}
Zhongye Jia and Hai Zhao. 2014.
\newblock A joint graph model for pinyin-to-chinese conversion with typo
  correction.
\newblock In \emph{Proceedings of the 52nd Annual Meeting of the Association
  for Computational Linguistics (Volume 1: Long Papers)}, pages 1512--1523.

\bibitem[{Jiao et~al.(2018)Jiao, Sun, and Sun}]{jiao2018LAC}
Zhenyu Jiao, Shuqi Sun, and Ke~Sun. 2018.
\newblock \href {https://arxiv.org/abs/1807.01882} {Chinese lexical analysis
  with deep bi-gru-crf network}.
\newblock \emph{arXiv preprint arXiv:1807.01882}.

\bibitem[{Lewis et~al.(2020)Lewis, Liu, Goyal, Ghazvininejad, Mohamed, Levy,
  Stoyanov, and Zettlemoyer}]{lewis-etal-2020-bart}
Mike Lewis, Yinhan Liu, Naman Goyal, Marjan Ghazvininejad, Abdelrahman Mohamed,
  Omer Levy, Veselin Stoyanov, and Luke Zettlemoyer. 2020.
\newblock \href {https://doi.org/10.18653/v1/2020.acl-main.703} {{BART}:
  Denoising sequence-to-sequence pre-training for natural language generation,
  translation, and comprehension}.
\newblock In \emph{Proceedings of the 58th Annual Meeting of the Association
  for Computational Linguistics}, pages 7871--7880, Online. Association for
  Computational Linguistics.

\bibitem[{Liu et~al.(2022)Liu, Song, Yue, Yang, Cai, Yu, and
  Sun}]{liu2022craspell}
Shulin Liu, Shengkang Song, Tianchi Yue, Tao Yang, Huihui Cai, TingHao Yu, and
  Shengli Sun. 2022.
\newblock Craspell: A contextual typo robust approach to improve chinese
  spelling correction.
\newblock In \emph{Findings of the Association for Computational Linguistics:
  ACL 2022}, pages 3008--3018.

\bibitem[{Liu et~al.(2021)Liu, Yang, Yue, Zhang, and Wang}]{liu2021plome}
Shulin Liu, Tao Yang, Tianchi Yue, Feng Zhang, and Di~Wang. 2021.
\newblock Plome: Pre-training with misspelled knowledge for chinese spelling
  correction.
\newblock In \emph{Proceedings of the 59th Annual Meeting of the Association
  for Computational Linguistics and the 11th International Joint Conference on
  Natural Language Processing (Volume 1: Long Papers)}, pages 2991--3000.

\bibitem[{Loshchilov and Hutter(2017)}]{loshchilov2017decoupled}
Ilya Loshchilov and Frank Hutter. 2017.
\newblock Decoupled weight decay regularization.
\newblock \emph{arXiv preprint arXiv:1711.05101}.

\bibitem[{Ma et~al.(2022)Ma, Li, Sun, Zhou, Huang, Zhang, Yangning, Liu, Li,
  Cao, Zheng, and Shen}]{ma-etal-2022-linguistic}
Shirong Ma, Yinghui Li, Rongyi Sun, Qingyu Zhou, Shulin Huang, Ding Zhang,
  Li~Yangning, Ruiyang Liu, Zhongli Li, Yunbo Cao, Haitao Zheng, and Ying Shen.
  2022.
\newblock \href {https://aclanthology.org/2022.findings-emnlp.40} {Linguistic
  rules-based corpus generation for native {C}hinese grammatical error
  correction}.
\newblock In \emph{Findings of the Association for Computational Linguistics:
  EMNLP 2022}, pages 576--589, Abu Dhabi, United Arab Emirates. Association for
  Computational Linguistics.

\bibitem[{Rao et~al.(2020)Rao, Yang, and Zhang}]{rao2020overview}
Gaoqi Rao, Erhong Yang, and Baolin Zhang. 2020.
\newblock Overview of nlptea-2020 shared task for chinese grammatical error
  diagnosis.
\newblock In \emph{Proceedings of the 6th Workshop on Natural Language
  Processing Techniques for Educational Applications}, pages 25--35.

\bibitem[{Tan et~al.(2022)Tan, Dai, Tang, Feng, Huang, Jiang, Li, and
  Shi}]{tan2022exploring}
Minghuan Tan, Yong Dai, Duyu Tang, Zhangyin Feng, Guoping Huang, Jing Jiang,
  Jiwei Li, and Shuming Shi. 2022.
\newblock Exploring and adapting chinese gpt to pinyin input method.
\newblock In \emph{Proceedings of the 60th Annual Meeting of the Association
  for Computational Linguistics (Volume 1: Long Papers)}, pages 1899--1909.

\bibitem[{Tseng et~al.(2015)Tseng, Lee, Chang, and
  Chen}]{YuenHsienTseng2015IntroductionTS}
Yuen-Hsien Tseng, Lung-Hao Lee, Li-Ping Chang, and Hsin-Hsi Chen. 2015.
\newblock Introduction to sighan 2015 bake-off for chinese spelling check.
\newblock \emph{CIPS-SIGHAN Joint Conference on Chinese Language Processing}.

\bibitem[{Wang et~al.(2022)Wang, Duan, Wu, Che, Chen, and
  Hu}]{wang-etal-2022-cctc}
Baoxin Wang, Xingyi Duan, Dayong Wu, Wanxiang Che, Zhigang Chen, and Guoping
  Hu. 2022.
\newblock \href {https://aclanthology.org/2022.coling-1.294} {{CCTC}: A
  cross-sentence {C}hinese text correction dataset for native speakers}.
\newblock In \emph{Proceedings of the 29th International Conference on
  Computational Linguistics}, pages 3331--3341, Gyeongju, Republic of Korea.
  International Committee on Computational Linguistics.

\bibitem[{Wang et~al.(2018)Wang, Song, Li, Han, and Zhang}]{wang2018hybrid}
Dingmin Wang, Yan Song, Jing Li, Jialong Han, and Haisong Zhang. 2018.
\newblock A hybrid approach to automatic corpus generation for chinese spelling
  check.
\newblock In \emph{Proceedings of the 2018 Conference on Empirical Methods in
  Natural Language Processing}, pages 2517--2527.

\bibitem[{Wang et~al.(2020)Wang, Wang, Gong, Wang, Hu, Duan, Shen, Yue, Fu, Wu
  et~al.}]{wang2020combining}
Shaolei Wang, Baoxin Wang, Jiefu Gong, Zhongyuan Wang, Xiao Hu, Xingyi Duan,
  Zizhuo Shen, Gang Yue, Ruiji Fu, Dayong Wu, et~al. 2020.
\newblock Combining resnet and transformer for chinese grammatical error
  diagnosis.
\newblock In \emph{Proceedings of the 6th Workshop on Natural Language
  Processing Techniques for Educational Applications}, pages 36--43.

\bibitem[{Wu et~al.(2013)Wu, Liu, and Lee}]{ShihHungWu2013ChineseSC}
Shih-Hung Wu, Chao-Lin Liu, and Lung-Hao Lee. 2013.
\newblock Chinese spelling check evaluation at sighan bake-off 2013.
\newblock \emph{CIPS-SIGHAN Joint Conference on Chinese Language Processing}.

\bibitem[{Xu et~al.(2022)Xu, Wu, Peng, Fu, and Cai}]{xu-etal-2022-fcgec}
Lvxiaowei Xu, Jianwang Wu, Jiawei Peng, Jiayu Fu, and Ming Cai. 2022.
\newblock \href {https://aclanthology.org/2022.findings-emnlp.137} {{FCGEC}:
  Fine-grained corpus for {C}hinese grammatical error correction}.
\newblock In \emph{Findings of the Association for Computational Linguistics:
  EMNLP 2022}, pages 1900--1918, Abu Dhabi, United Arab Emirates. Association
  for Computational Linguistics.

\bibitem[{Yu et~al.(2014)Yu, Lee, Tseng, and Chen}]{LiangChihYu2014OverviewOS}
Liang-Chih Yu, Lung-Hao Lee, Yuen-Hsien Tseng, and Hsin-Hsi Chen. 2014.
\newblock Overview of sighan 2014 bake-off for chinese spelling check.
\newblock \emph{CIPS-SIGHAN Joint Conference on Chinese Language Processing}.

\bibitem[{Zhang et~al.(2020)Zhang, Huang, Liu, and Li}]{zhang2020spelling}
Shaohua Zhang, Haoran Huang, Jicong Liu, and Hang Li. 2020.
\newblock Spelling error correction with soft-masked bert.
\newblock In \emph{Proceedings of the 58th Annual Meeting of the Association
  for Computational Linguistics}, pages 882--890.

\bibitem[{Zhao et~al.(2022)Zhao, Wang, Wu, Che, Chen, and
  Wang}]{zhao2022overview}
Honghong Zhao, Baoxin Wang, Dayong Wu, Wanxiang Che, Zhigang Chen, and Shijin
  Wang. 2022.
\newblock Overview of ctc 2021: Chinese text correction for native speakers.
\newblock \emph{arXiv preprint arXiv:2208.05681}.

\bibitem[{Zhu et~al.(2022)Zhu, Ying, Zhang, and Mao}]{zhu2022mdcspell}
Chenxi Zhu, Ziqiang Ying, Boyu Zhang, and Feng Mao. 2022.
\newblock Mdcspell: A multi-task detector-corrector framework for chinese
  spelling correction.
\newblock In \emph{Findings of the Association for Computational Linguistics:
  ACL 2022}, pages 1244--1253.

\end{thebibliography}
\bibliographystyle{acl_natbib}

\appendix

\clearpage
\section{Pseudo Data Analysis}

\subsection{Impact of LM Post-Filtering}

\begin{table}
  \small
  \centering
  \def\arraystretch{1.2}
  \begin{tabular}{cccc}
  \hline
  LM threshold ($\delta$) & Precision & Recall & F1 \\
  \hline
  w/o & 41.17 & 40.66 & 40.91 \\ 
  -20\% & 44.52 & \textbf{49.01} & 46.66 \\ 
  0\% &  49.69 & 44.07 & \textbf{46.71} \\ 
  20\% &  50.64 & 26.46 & 34.76 \\ 
  50\% & \textbf{57.52} & 9.38 & 16.12 \\ 
  \hline
\end{tabular}
\caption{The correction results (\%) at character level for pseudo data with different LM filtering strategies.}
\label{Tab:ngram}
\end{table}

In this section, we investigate the influence of language model (LM) post-filtering, which constitutes the final stage of our proposed pseudo-data construction method. We extract accurate sentences from the Wang271K dataset \citep{wang2018hybrid} and generate pseudo-data using IME, incorporating various LM filtering strategies. We choose the basic BERT model to conduct the experiment and train the model only on the pseudo data to clearly distinguish the differences.

As demonstrated in Table \ref{Tab:ngram}, the lack of LM filtering results in the introduction of undesired noise. For example, the generated pseudo-data may consist of entirely accurate sentences. In contrast, when the threshold is excessively low (even below 0), the generated errors become more complex, leading to high recall but poor precision. Conversely, if the threshold is set too high, the generated errors tend to be relatively simple, resulting in better precision but lower recall. Therefore, LM filtering is necessary, and selecting an appropriate threshold is also very important.

\begin{figure*}[htbp]
	\centering
	\includegraphics[width=0.95\textwidth]{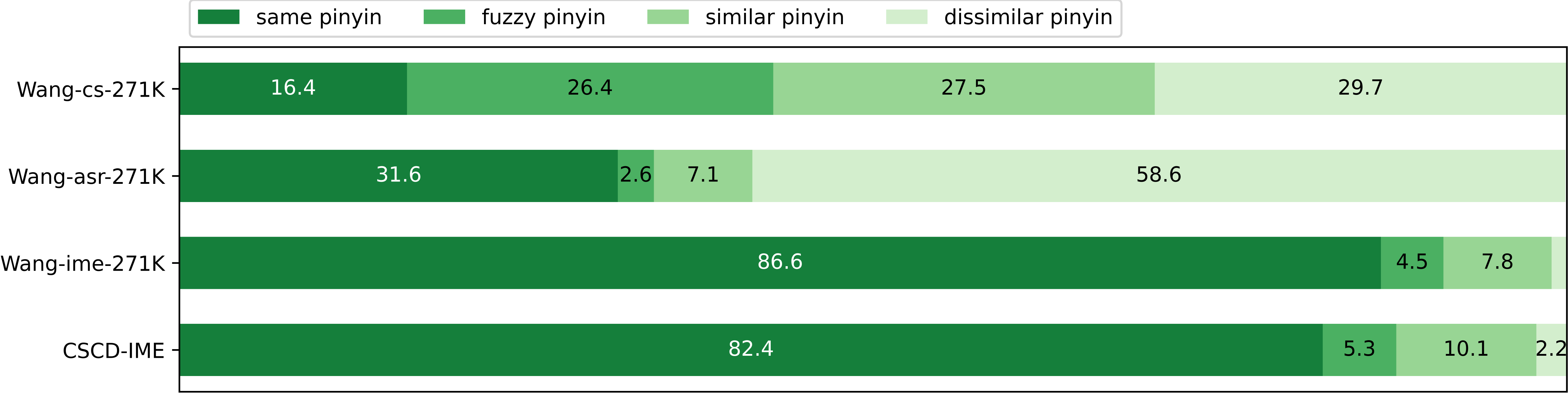}
	\includegraphics[width=0.95\textwidth]{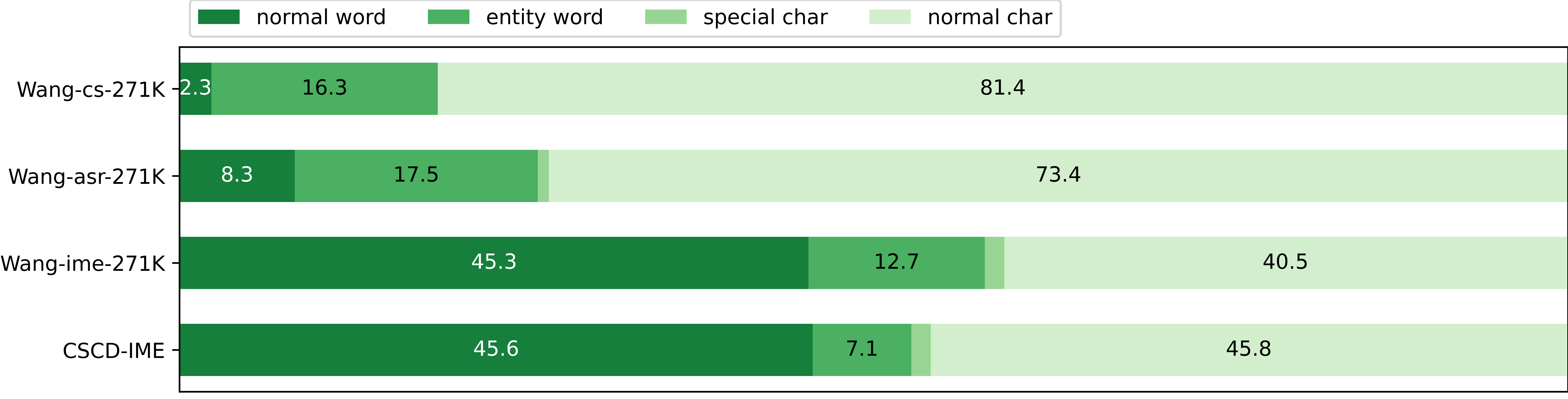}
	\caption{The comparison of error distribution (\%) at phonetic level (above) and semantic level (below).}
	\label{fig:analysis-pseudo}
\end{figure*}

\subsection{Error Distribution}
As illustrated in Figure \ref{fig:analysis-pseudo}, we analyze the error distribution of pseudo-data generated by various methods at both phonetic and semantic levels. It is clear that our pseudo-data construction method demonstrates the highest consistency with the CSCD-NS dataset, suggesting that our approach closely resembles real input scenarios. In contrast, the confusion set-based method and the ASR-based method exhibit a significant deviation from the actual error distribution.

\begin{CJK*}{UTF8}{gbsn}
  {
    \begin{table}
    \small
    \centering
    \def\arraystretch{1.2}
    \begin{tabular}{cl}
    \hline
    \textit{translation} & simple, fashionable and moderate style \\
    \textit{origin} & 简约时尚的风格适中的 \\
    \textit{CS} & 简约时尚的风格\textcolor{red}{誓}中的\\
    \textit{ASR} & 简约时尚的风格\textcolor{red}{是}中的 \\
    \textit{IME} & 简约时尚的风格\textcolor{red}{始终}的 \\
    \hline
    \textit{translation} & and the regulation is not perfect \\
    \textit{origin} & 且监管也不完善 \\
    \textit{CS} & 且监管也不\textcolor{red}{碗}善 \\
    \textit{ASR} & \textcolor{red}{其}监管也不完善 \\
    \textit{IME} & 且监管也不\textcolor{red}{玩}善 \\
    \hline
  \end{tabular}
  \caption{The pseudo data generated based on confusion set (CS), ASR, and IME.}
  \label{Tab:pseudo-example}
  \end{table}
  }
  \end{CJK*}
  
\subsection{Case Study}
We sample some examples in Table \ref{Tab:pseudo-example}. It can be observed that the confusion set-based method is capable of producing similar phonetic errors; however, these errors are entirely out of context and can not accurately represent the real input scenario. The ASR-based method performs better but primarily generates character-level errors. Moreover, since the ASR-based method lacks an LM filtering module, the generated noise may occasionally be correct, as demonstrated by the third case in Table \ref{Tab:pseudo-example}. In contrast, our method can effectively generate high-quality pseudo data, encompassing both word-level and character-level errors.

\begin{table}
  \small
  \centering
  \def\arraystretch{1.2}
  \begin{tabular}{ll}
  \hline
  Configurations & Values \\
  \hline
  PLM & bert-base-chinese \citep{devlin-etal-2019-bert} \tablefootnote{https://huggingface.co/bert-base-chinese} \\
  devices & 1 Nvidia A100 GPU (40GB) \\
  framework & PyTorch Lightning 1.3.8 \tablefootnote{https://www.pytorchlightning.ai/} \\
  optimizer & AdamW \citep{loshchilov2017decoupled} \\
  learning rate & 1e-4 \\
  sequence length & 512 \\
  batch size & 128 \\
  epochs & 10 \\
  dropout & 0.1 \\
  \multirow{2}{*}{model size} & BERT: 102 M \\
  & SM BERT: 123 M  \\
  \multirow{2}{*}{training speed} & BERT: \textasciitilde10 batches/s \\
  & SM BERT: \textasciitilde7 batches/s \\
  metric for best \tablefootnote{The metric used to save the best model} & loss \\
  \hline
\end{tabular}
  \caption{Configurations of BERT and SM BERT.}
\label{Tab:bert}
\end{table}

\begin{table}
  \small
  \centering
  \def\arraystretch{1.2}
  \begin{tabular}{ll}
  \hline
  Configurations & Values \\
  \hline
  PLM & PLOME pre-trained model \tablefootnote{https://share.weiyun.com/OREEY0H3} \\
  devices & 1 Nvidia V100 GPU (32GB) \\
  framework & Tensorflow 1.14 \tablefootnote{https://www.tensorflow.org/} \\
  optimizer & AdamW \citep{loshchilov2017decoupled} \\
  learning rate & 5e-5 \\
  sequence length & 180 \\
  batch size & 32 \\
  epochs & 10 \\
  dropout & 0.1 \\
  model size & 123 M \\
  training speed & \textasciitilde2.12 batches/s\\
  metric for best & F1-score of correction at character level \\
  \hline
\end{tabular}
  \caption{Configurations of PLOME}
\label{Tab:plome}
\end{table}

\section{Experimental Details}
In this section, we provide comprehensive descriptions of the experimental procedures and parameter settings for each model. 

Note that for each experiment, we select the best checkpoint based on the development set and evaluate its performance on the test set. We carry out three trials for each experiment and report the average results in the paper. The total training time is contingent upon the size of the training data and can be estimated based on the training speed.

\subsection{BERT-like Models}
Since there is no official implementation for BERT and SM BERT, we follow a widely-used open-source version\footnote{https://github.com/gitabtion/BertBasedCorrectionModels}. For PLOME, we directly utilize the official code\footnote{https://github.com/liushulinle/PLOME}. We adhere to the default hyperparameters, and the detailed configurations for these three models can be found in Table \ref{Tab:bert} and Table \ref{Tab:plome}.

\subsection{BART}
\begin{table}
  \small
  \centering
  \def\arraystretch{1.2}
  \begin{tabular}{ll}
  \hline
  Configurations & Values \\
  \hline
  PLM & fnlp/bart-large-chinese \tablefootnote{https://huggingface.co/fnlp/bart-large-chinese} \\
  devices & 8 Nvidia A100 GPU (40GB) \\
  framework & transformers 4.29.1 \tablefootnote{https://github.com/huggingface/transformers} \\
  optimizer & AdamW \citep{loshchilov2017decoupled} \\
  learning rate & 5e-5 \\
  sequence length & 512  \\
  batch size & 256 \\
  epochs & 10 \\
  dropout & 0.1 \\
  model size & 407 M \\
  training speed & \textasciitilde3.5 batches/s\\
  metric for best & loss \\
  \hline
  input & \{origin sentence\} \\
  \hline
  output & \{correct sentence \} \\
  \hline
\end{tabular}
  \caption{Configurations of BART}
\label{Tab:bart}
\end{table}

We choose the Chinese BART-large model as the base model and fine-tune it for the CSC task by treating it as a sequence-to-sequence task. The model takes the original sentence as input and produces the correct sentence as output. The decoding method employed is beam search with a beam size of 4. The specific model configuration can be found in Table \ref{Tab:bart}.

\subsection{Baichuan2}
\begin{CJK*}{UTF8}{gbsn}
Baichuan2 \citep{baichuan2023baichuan2} is a powerful Chinese language model that includes two open-source models, Baichuan2-7B and Baichuan2-13B. The CSC task is modeled as an instruction tuning task, with the instruction being "纠正句子中的拼写错误" (correct the spelling errors in the following sentence). We use LoRA \citep{hu2021lora} to fine-tune the model. During the decoding stage, random sampling is not performed, and the beam size is set to 1. Table \ref{Tab:chatglm} displays the specific configurations.

\begin{table}
  \small
  \centering
  \def\arraystretch{1.2}
  \begin{tabular}{ll}
  \hline
  Configurations & Values \\
  \hline
  PLM & Baichuan2 \tablefootnote{https://github.com/baichuan-inc/Baichuan2} \\
  devices & 8 Nvidia A100 GPU (40GB) \\
  framework & transformers 4.29.1 \tablefootnote{https://github.com/huggingface/transformers} \\
  optimizer & AdamW \\
  lora rank &  8 \\
  learning rate & 1e-4 \\
  sequence length & 512  \\
  batch size & 128   \\
  epochs & 10 \\
  dropout & 0.1 \\
  \multirow{2}{*}{model size} & Baichuan2-7B: 7.5 M \\
  & Baichuan2-13B: 13.9 M  \\
  \multirow{2}{*}{training speed} & Baichuan2-7B: \textasciitilde3.0 s/batch \\
  & Baichuan2-13B: \textasciitilde4.4 s/batch \\
  metric for best & loss \\
  \hline
  input & \tabincell{l}{Instrction: 纠正句子中的拼写错误 \\
  Input: \{origin sentence\} \\
  Output: } \\
  \hline
  output & \{correct sentence \} \\
  \hline
\end{tabular}
  \caption{Configurations of Baichuan2}
\label{Tab:chatglm}
\end{table}

\subsection{ChatGPT and GPT4}

\begin{table*}
  \small
  \centering
  \def\arraystretch{1.5}
  \begin{tabular}{ll}
  \hline
  \multicolumn{2}{c}{prompt 1} \\
  \hline
  instruction & 修正句子中的拼写错误，修正结果需要与原文长度相等，发音相近 \\
  \cline{2-2}
  10 examples & \tabincell{l}{ 比特币价格从15美元飚升到266美元 \ $\Rightarrow$ \  比特币价格从15美元飙升到266美元 \\
  ... \\
  其中，企业成为职务专利申请的主力军 \ $\Rightarrow$ \ 其中，企业成为职务专利申请的主力军 } \\
  \cline{2-2}
  test case & 让农民工流血、流汗不在流泪 \ $\Rightarrow$ \  \\
  \hline
  \hline
  \multicolumn{2}{c}{prompt 2} \\
  \hline
  instruction & 修正拼写错误，修正结果与原文需要长度相等，且发音尽可能相近 \\
  \cline{2-2}
  10 examples & \tabincell{l}{ 修正前:\ 比特币价格从15美元飚升到266美元 \\ 
  修正后: \ 比特币价格从15美元飙升到266美元 \\
  ... \\
  修正前:\ 其中，企业成为职务专利申请的主力军 \\ 
  修正后:\ 其中，企业成为职务专利申请的主力军 } \\
  \cline{2-2}
  test case & \tabincell{l}{修正前:\ 让农民工流血、流汗不在流泪 \\
  修正后: 
  } \\
  \hline
  \hline
  \multicolumn{2}{c}{prompt 3} \\
  \hline
  instruction & \tabincell{l}{Instruction: Correct spelling errors in the sentence, adhering to the following two requirements: \\
  (1) The corrected output should maintain the same character length as the original text. \\
  (2) The pinyin of the corrected character and the original character should be identical, or the edit distance \\ 
 should be as minimal as possible.}\\
  \cline{2-2}
  10 examples & \tabincell{l}{ Input:\ 比特币价格从15美元飚升到266美元 \\ 
  Output: \ 比特币价格从15美元飙升到266美元 \\
  ... \\
  Input:\ 其中，企业成为职务专利申请的主力军 \\ 
  Output:\ 其中，企业成为职务专利申请的主力军 } \\
  \cline{2-2}
  test case & \tabincell{l}{Input:\ 让农民工流血、流汗不在流泪 \\
  Output: 
  } \\
  \hline
\end{tabular}
  \caption{Three prompt templates designed to call ChatGPT/GPT4 for the CSC task.}
\label{Tab:chatgpt_prompt}
\end{table*}

\begin{table*}
  \small 
  \centering
  \bgroup
  \setlength\tabcolsep{0.6em}
  \def\arraystretch{1.2}
  \begin{tabular}{c|ccc|ccc|ccc|ccc}
  \hline
  \multirow{3}{*}{Settings} & \multicolumn{6}{c|}{Sentence level} & \multicolumn{6}{c}{Character level} \\
  \cline{2-13}
  & \multicolumn{3}{|c|}{Detection} & \multicolumn{3}{|c|}{Correction} & \multicolumn{3}{|c|}{Detection} & \multicolumn{3}{|c}{Correction} \\
  \cline{2-13}
  & P & R & F1 & P & R & F1 & P & R & F1 & P & R & F1 \\
  \hline
  prompt 1 (run1) & 52.92 & 51.13 & 52.01 & 48.70 & 47.05 & 47.86 & 54.14 & 57.91 & 55.96 & 48.56 & 51.94 & 50.19 \\
  prompt 1 (run2) & 53.61 & 50.22 & 51.86 & 49.40 & 46.27 & 47.78 & 54.08 & 56.28 & 55.16 & 48.84 & 50.83 & 49.82 \\
  prompt 1 (run3) & 53.85 & 50.61 & 52.18 & 49.75 & 46.75 & 48.20 & 54.73 & 56.92 & 55.80 & 49.30 & 51.27 & 50.26 \\
  \hline
  prompt 2 (run1) & 55.52 & 48.83 & 51.96 & 50.94 & 44.80 & 47.67 & 55.08 & 54.86 & 54.97 & 49.25 & 49.05 & 49.15 \\ 
  prompt 2 (run2) & 55.43 & 49.61 & 52.36 & 50.82 & 45.49 & 48.01 & 55.48 & 55.65 & 55.56 & 49.72 & 49.88 & 49.80 \\
  prompt 2 (run3) & 55.91 & 50.22 & 52.91 & 51.76 & 46.49 & 48.98 & 55.56 & 56.72 & 56.13 & 50.33 & 51.38 & 50.85 \\
  \hline
  prompt 3 (run1) & 59.56 & 47.27 & 52.71 & 55.25 & 43.84 & 48.89 & 61.16 & 51.11 & 55.69 & 55.49 & 46.36 & 50.52 \\
  prompt 3 (run2) & 58.29 & 45.88 & 51.35 & 54.88 & 43.19 & 48.34 & 60.62 & 49.84 & 54.71 & 55.67 & 45.77 & 50.24 \\
  prompt 3 (run3) & 59.85 & 47.83 & 53.17 & 55.56 & 44.41 & 49.36 & 61.29 & 51.70 & 56.09 & 56.00 & 47.23 & 51.24 \\
  \hline
\end{tabular}
\egroup
\caption{The performance (\%) of ChatGPT with different prompts on CSCD-NS.}
\label{Tab:chatgpt_result}
\end{table*}

We tested ChatGPT and GPT4 through OpenAI's API on November 26, 2023, and the model id for ChatGPT is \textit{gpt-3.5-turbo-1106} and GPT4 is \textit{gpt-4-1106-preview}.
We set the temperature to 0 to reduce the influence of random sampling. As illustrated in Table \ref{Tab:chatgpt_prompt}, we devise three prompt templates, each comprising a task description, 10 examples, and a test sentence. These 10 examples encompass 5 positive instances (sentences containing spelling errors) and 5 negative instances (sentences without spelling errors), all of which are randomly chosen from the training set. As shown in Table \ref{Tab:chatgpt_result}, utilizing the same prompt template with varying example samples exerted a negligible effect on the outcomes. Likewise, employing different prompt templates also has a minor impact on the results. Given that the outcomes obtained using "prompt 3" are slightly better, we present the average results derived from "prompt 3" in our paper.
\end{CJK*}

\end{document}